# Software Agents: Completing Patterns and Constructing User Interfaces

**Jeffrey C. Schlimmer**                                    SCHLIMMER@EECS.WSU.EDU
**Leonard A. Hermens**                                    LHERMENS@EECS.WSU.EDU
*School of Electrical Engineering & Computer Science,*
*Washington State University, Pullman, WA 99164-2752, U.S.A.*

## Abstract

To support the goal of allowing users to record and retrieve information, this paper describes an interactive note-taking system for pen-based computers with two distinctive features. First, it actively predicts what the user is going to write. Second, it automatically constructs a custom, button-box user interface on request. The system is an example of a learning-apprentice software-agent. A machine learning component characterizes the syntax and semantics of the user's information. A performance system uses this learned information to generate completion strings and construct a user interface.

## 1. Introduction and Motivation

People like to record information for later consultation. For many, the media of choice is paper. It is easy to use, inexpensive, and durable. To its disadvantage, paper records do not scale well. As the amount of information grows, retrieval becomes inefficient, physical storage becomes excessive, and duplication and distribution become expensive. Digital media offers better scaling capabilities. With indexing and sub-linear algorithms, retrieval is efficient; using high density devices, storage space is minimal; and with electronic storage and high-speed networks, duplication and distribution is fast and inexpensive. It is clear that our computing environments are evolving as several vendors are beginning to market inexpensive, hand-held, highly portable computers that can convert handwriting into text. We view this as the start of a new paradigm shift in how traditional digital information will be gathered and used. One obvious change is that these computers embrace the paper metaphor, eliminating the *need* for typing. It is in this paradigm that our research is inspired, and one of our primary goals is to combine the best of both worlds by making digital media as convenient as paper.

This document describes an interactive note-taking software system for computers with pen-based input devices. Our software has two distinctive features: first, it actively predicts what the user is going to write and provides a default that the user may select; second, the software automatically constructs a graphical interface at the user's request. The purpose of these features is to speed up information entry and reduce user errors. Viewed in a larger context, the interactive note-taking system is a type of self-customizing software.

To clarify this notion, consider a pair of dimensions for characterizing software. As Figure 1 depicts, one dimension is task specificity. Software that addresses a generic task (e.g., a spreadsheet) lies between task independent software (e.g., a compiler) and task specific software (e.g., a particular company's accounting software). Another dimension is the amount of user customization required to make the software useful. Task generic software lies between the two extremes, requiring modest programming in a specialized language. Self-customizing software uses machine learning techniques to automatically customize task generic software to a specific user. Because the software learns to assist the user by watching them complete tasks, the software is also a learning apprentice. Similarly, because the user does not explicitly program the defaults or the user interface for the note taking system, it is a type of software agent. Agents are a new user interface paradigm that free the user from having to explicitly command the computer. The user can record information directly and in a free-form manner. Behind the interface, the software is acting on behalf of the user, helping to capture and organize the information.

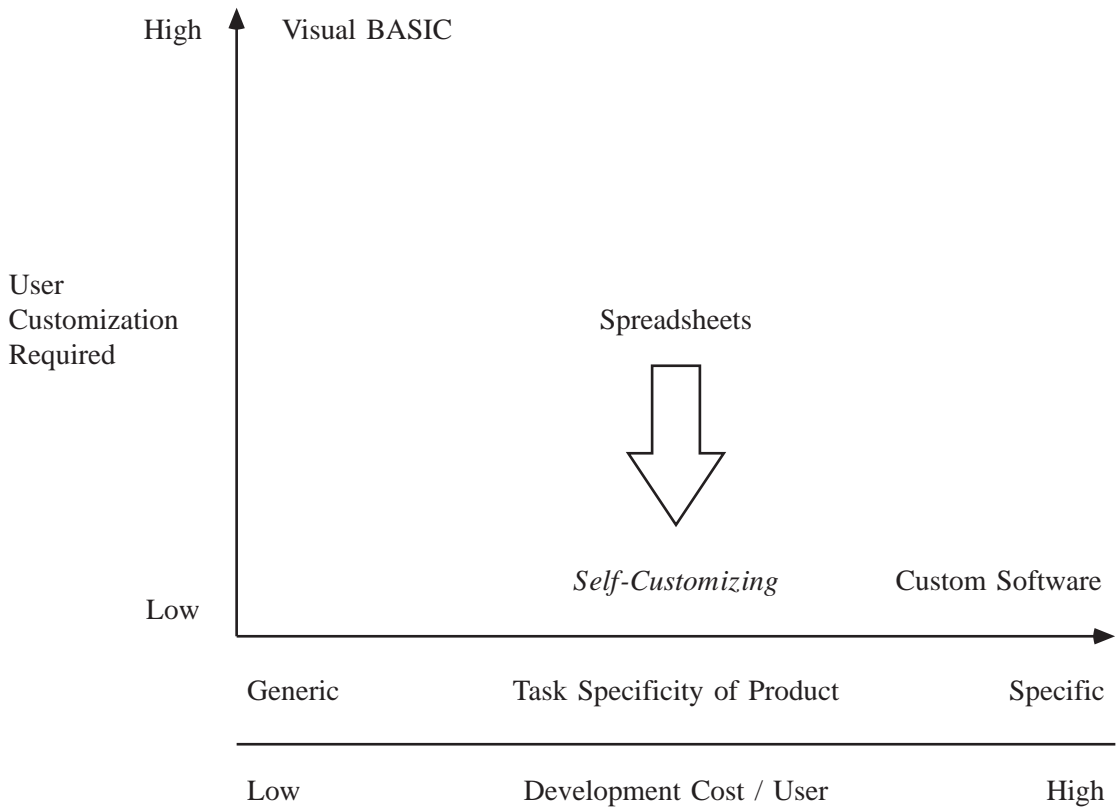

Figure 1: Continuum of software development depicting the traditional trade-off between the development cost per user and the amount of user customization required. Self-customizing software eliminates the need for user customization by starting with partially-specified software and applying machine learning methods to complete any remaining customization.

Next we will introduce the performance component of the note-taking software in more detail, then describe the representations and algorithms used by the learning methods. We also present empirical results, comparing the performance of seven alternate methods on nine realistic note-taking domains, and finally, we describe related research and identify some of the system's limitations.

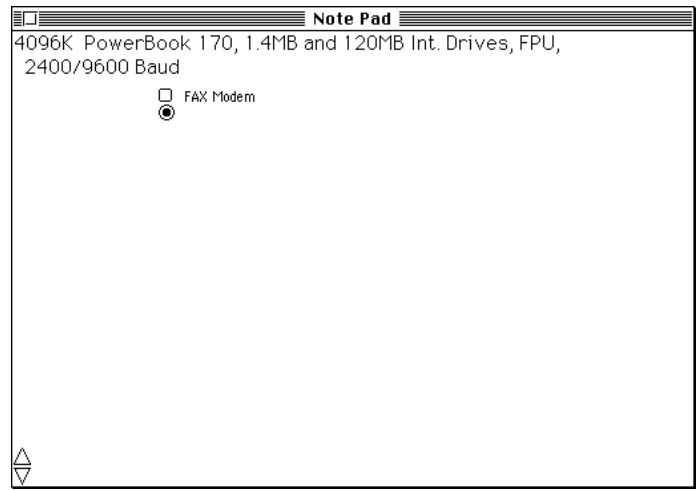

Figure 2: Screen snapshot of the note-taking software in contextual prompting mode for a PowerBook note. The two triangles in the lower left are scroller buttons.

## 2. Performance Task

The primary function of the note-taking software is to improve the user's speed and accuracy as they enter notes about various domains of interest. A *note* is a short sequence of descriptive terms that describe a single object of interest. Example 1 shows a note describing a particular personal computer (recorded by the first author from a Usenet newsgroup during 1992):

```
4096K PowerBook 170, 1.4MB and 40MB Int. Drives, 2400/9600 Baud FAX Modem
```
                                                                    (Example 1)

Example 2 is a note describing a fabric pattern (recorded by the first author's wife):

```
          Butterick 3611 Size 10 dress, top                    (Example 2)
```

Tables 5 through 11 later in the paper list sample notes drawn from seven other domains. The user may enter notes from different domains at their convenience and may use whatever syntactic style comes naturally.

From the user's point of view, the software operates in one of two modes: a *contextual* prompting mode, and an interactive graphical interface mode. In the first mode, the software continuously predicts a likely completion as the user writes out a note. It offers this as a default for the user. The location and presentation of this default must balance conflicting requirements to be convenient yet unobtrusive. For example, the hand should not hide the indicated default while the user is writing. Our solution is to have a small, colored completion button follow to the left and below where the user is writing. In this location, it is visible to either right- or left-handed people as they write out notes. The user can reposition the button to another location if they prefer. The default text is displayed to the immediate right of this button in a smaller font. The completion button is green; the text is black. The completion button saturation ranges from 1 (appearing green), when the software is highly confident of the predicted value, to 0 (appearing white), when the software lacks confidence. The button has a light gray frame, so it is visible even when the software has no prediction. Figure 2 portrays a screen snapshot of the software operating in the contextual prompting mode for a PowerBook note.

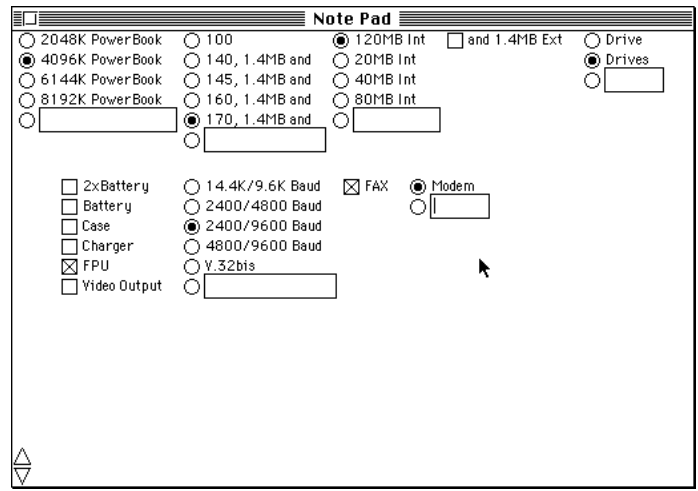

Figure 3: Screen snapshot of the note-taking software in button-box mode for a PowerBook note.

The software's second mode presents an interactive graphical interface. Instead of requiring the user to write out the text of a note, the software presents a radio-button and check-box interface (what we call a *button-box interface*). With this, the user may select from text fragments, portions of notes called *descriptive terms*, by tapping on radio-buttons or check-boxes with the pen interface device. Each selection from the button-box interface is added to the current note. Intuitively, check boxes are generated to depict optional descriptive terms, whereas radio-button panels are generated to depict alternate, exclusive descriptive terms. For user convenience, the radio-buttons are clustered into panels and are sorted alphabetically in ascending order from top to bottom. To allow the user to add new descriptive terms to a button-box panel, an additional blank button is included at the bottom of each. When the user selects a radio button item, the graphical interface is expanded to depict additional choices corresponding to descriptive terms that follow syntactically. The software indicates its predictions by preselecting the corresponding buttons and highlighting them in green. The user may easily override the default selection by tapping the desired button. Figure 3 portrays a screen snapshot of the software operating in the interactive graphical interface mode for a PowerBook note.

The software is in prompting mode when a user begins to write a note. If the learned syntax for the domain of the note is sufficiently mature (see Section 6, Constructing a Button-Box Interface), then the software can switch into the button-box mode. To indicate this to the user, a mode switch depicted as a radio button is presented for the user's notice. A convenient and unobtrusive location for this switch is just below the completion button. In keeping with the color theme, the mode switch also has a green hue. If the user taps this switch, the written text is removed, and the appropriate radio buttons and check boxes are inserted. The system automatically selects buttons that match the user-written text. As the user makes additional selections, the interface expands to include additional buttons. When the user finishes a note, in either mode, the software returns to prompting mode in anticipation of another note.[1] Because the interface is constructed from a learned syntax, as the software refines its representation of the domains of the notes, the button-box interface also improves. On-line Appendix 1 is a demonstration of the system's operation in each of its two modes.

## 3. Learning a Syntax

To implement the two modes of the note taking software, the system internally learns two structures. To characterize the syntax of user's notes, it learns finite-state machines (FSMs). To generate predictions, it learns decision tree classifiers situated at states within the FSMs. In order to construct a graphical user interface, the system converts a FSM into a set of buttons. This section describes the representation and method for learning FSMs. The next section discusses learning of the embedded classifiers.

### 3.1 Tokenization

Prior to learning a finite-state machine, the user's note must first be converted into a sequence of tokens. Useful tokenizers can be domain independent. However, handcrafted domain-specific tokenizers lead to more useful representations. The generic tokenizer used for the results reported here uses normal punctuation, whitespace, and alpha-numeric character boundaries as token delimiters. For example, our generic tokenizer splits the sample PowerBook note in Example 1 into the following 16 tokens:

```
:NULL
"4096"
" K"
" PowerBook"
" 170"
", 1.4"
"MB"
" and"
" 40"
"MB"
" Int."
" Drives"
", 2400/9600"
" Baud"
" FAX"
" Modem".
```

The token `:NULL` is prepended by the tokenizer. This convention simplifies the code for constructing a FSM.

### 3.2 Learning a Finite-State Machine

Deterministic finite-state machines (FSMs) are one candidate approach for describing the syntax of a user's notes because they are well understood and relatively expressive. Moreover, Angluin (1982) and Berwick and Pilato (1987) present a straightforward algorithm for learning a specific subclass of FSMs called k-reversible FSMs. The algorithm is incremental

---

1. Of the functionality described here, our prototype implements all but the transition from button-box to contextual prompting. The mechanism for such a transition is machine dependent and is not germane to this research.

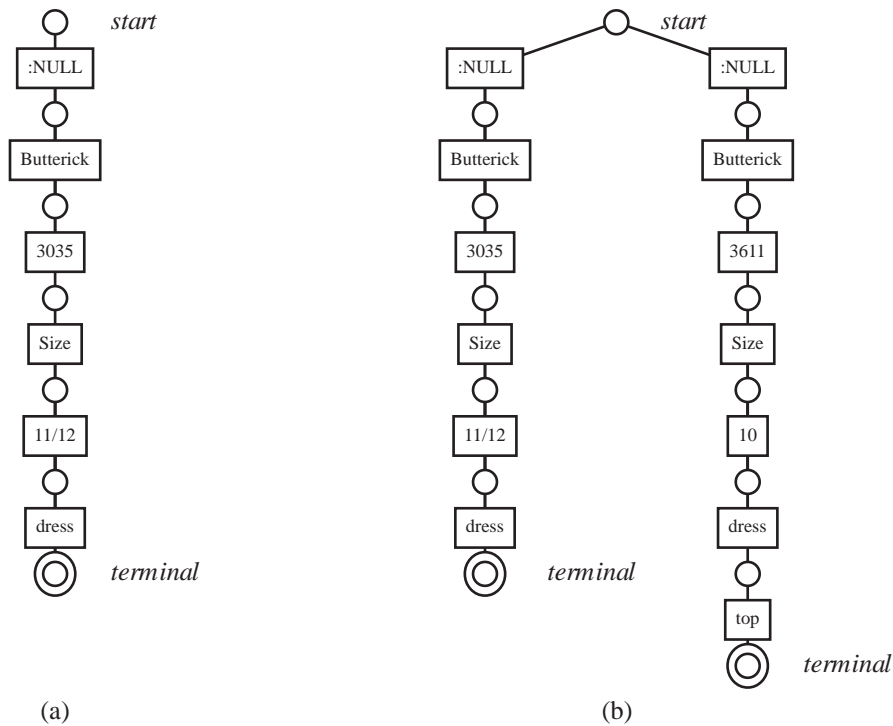

(a)                    (b)

Figure 4: (a) Degenerate finite-state machine after processing a single fabric pattern note, and (b) prefix tree finite-state machine after adding a second fabric pattern note (cf. Example 2).

and does not suffer from presentation order effects. Berwick and Pilato define a k-reversible FSM as:

> "A regular language is *k-reversible*, where *k* is a non-negative integer, if whenever two prefixes *whose last k words [tokens] match* have a tail in common, then the two prefixes have all tails in common. In other words, a deterministic finite-state automaton (DFA) [FSM] is *k*-reversible if it is deterministic with lookahead *k* when its sets of initial and final states are swapped and all of its arcs [transitions] are reversed."

Given a list of tokens, the k-reversible FSM algorithm first constructs a prefix tree, where all token sequences with common k-leaders share a k-length path through the FSM. For example, Figure 4a depicts a simple FSM constructed for a single fabric pattern note. The text of the user's note was converted into a sequence of tokens. Then a transition was created for each token and a sequence of states was created to link them together. One state serves as the initial state, and another indicates the completion of the sequence. For convenience, this latter, terminal state is depicted with a double circle. If the FSM is able to find a transition for each token in the sequence, and it arrives at the terminal state, then the FSM accepts the token sequence as an instance of the language it defines. Figure 4b depicts the same FSM after another path has been added corresponding to a second fabric pattern note (Example 2). Now the FSM will accept either note if expressed as a sequence of tokens. This FSM is a trivial prefix tree because only the first state is shared between the two paths.

---

A k-leader is defined as a path of length k that accepts in the given state.
Merge any two states if either of the following is true:

1. Another state transitions to both states on the same token; or
   (This enforces determinism.)
2. Both states have a common k-leader and
   a. Both states are accepting states, or
   b. Both states transition to a common state via the same token.

---

Table 1: FSM state merging rules from (Angluin, 1982).

A prefix tree is minimal for observed token sequences, but it may not be general enough for use in prediction. (The prefix tree is, in essence, an expensive method for memorizing token sequences—which is not the desired result.) For the sake of prediction, it is desirable to have a FSM that can accept new, previously unseen combinations of tokens. The prefix tree automaton can be converted into a more general FSM by merging some of its states. A particular method for doing this converts a prefix tree into a k-reversible FSM via Angluin's (1982) algorithm. The algorithm merges states that have similar transitions, and it creates a FSM that accepts all token sequences in the prefix tree, as well as other candidate sequences. Table 1 lists the three rules for deciding when to merge a pair of states in a prefix tree to form a k-reversible FSM. In the special case where k equals zero, all states have a common k-leader, and Rule 2a ensures that there will be only one accepting state.

Because the rules in Table 1 must be applied to each pair of states in the FSM, and because each time a pair of states is merged the process must be repeated, the asymptotic complexity of the process is $O(n^3)$, where $n$ is the number of states in the FSM.

Applying these rules to the prefix tree in Figure 4b with k equal to zero results in a FSM depicted in Figure 5a. Notice that the first two states have been merged to make the FSM deterministic (Rule 1). The accepting states have also been merged in compliance with Rule 2a. The resulting FSM has fewer states but is not more general. It only accepts the two token sequences originally seen. Extending this example, Figure 5b illustrates the addition of a third fabric pattern note as a prefix tree path to the FSM. Reapplying the rules results in the FSM shown in Figure 6. The first two states have been merged as before through the action of the determinism Rule 1. Note that a pair of latter states have also been merged because they share a common zero-leader (true of all pairs of states) and because they transition to the common terminal state on the token `dress`.

Figure 7 depicts a more sophisticated result; it shows a learned zero-reversible FSM for notes about PowerBook computers. This example shows that the model number `100` is never followed by a specification for an internal floppy drive, but that other model numbers are. Any model may have an external floppy drive. Note that there is a single terminal state. Whitespace and punctuation have been eliminated for clarity in the figure.

The rules listed in Table 1 are generalization operators that allow the FSM to accept previously unobserved sequences. Whenever two or more states are merged into one, the FSM will accept more sequences than before if the new state is at the tail end of more transitions than one of the previous states and if the new state is at the head end of at least one transition. For example, the state just after State 1 in Figure 7 was merged from several previous states and generalizes memory sizes for PowerBook models. These rules comprise a heuristic bias and may be too conservative. For example, Figure 8 depicts a FSM for notes about fabric patterns. Many of the states prior to the accepting state could be usefully merged, but using only the rules listed in Table 1, many more notes will have to be processed before this happens. If the FSM in Figure 8 were rendered as a button-box interface, it would reflect little of the true structure of the domain of fabric patterns. Table 2 lists specializations of Rules 2a and 2b and an additional pair of rules we developed to make the FSM generalize more readily. Note that the parameter k has been set to zero in Rule 2 and to one in Rule 3. Effectively, two states are merged by Rules 3a or 2b' if they share an incoming or outgoing transition. Rule 3b is a Kleene rule that encourages the FSM to generalize the number of times a token may appear in a sequence. If one state has a transition to another, then merging them will result in a transition that loops from and to the newly merged state. Figure 9 depicts a FSM for notes about fabric patterns learned using all three generalization rules in Table 2. The resulting FSM accurately captures the syntax of the user's fabric pattern notes and correctly indicates the syntactically optional tokens that may appear at the end of note. When rendered as a button-box interface, it clearly depicts the user's syntax (as illustrated later by Figure 12). The added generalization rules may have only marginal effects on the system's ability to accurately predict a completion as the user writes out a note (as Table 14 below indicates). Their purpose is to improve the quality of the custom interface.

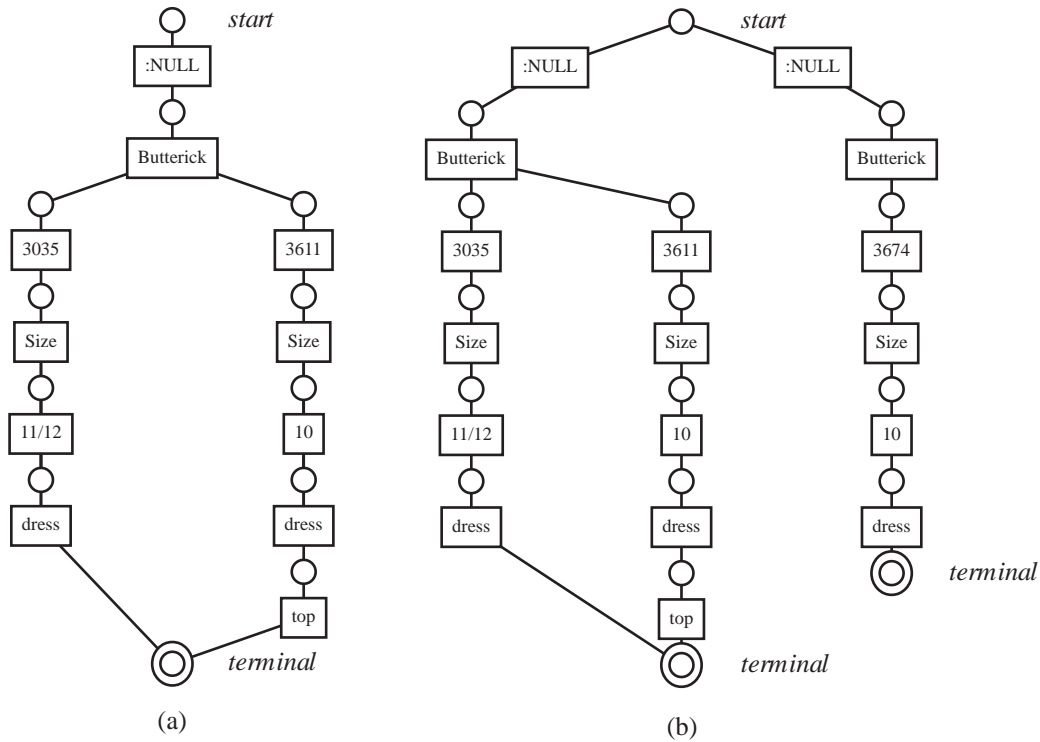

Figure 5: (a) Finite-state machine after processing two fabric pattern notes and applying state merging rules in Table 1, and (b) prefix tree finite-state machine after adding a third fabric pattern note.

Cohen (1988) uses an interesting alternative representation for learning a syntactic form. The goal in his work is to guide the generation of proof structures. Intuitively, the representation is a finite-state machine that accepts a tree rather than a sequence, and for this reason it is termed a tree automaton. Like the rules in Tables 1 and 2, tree automatons are generalized

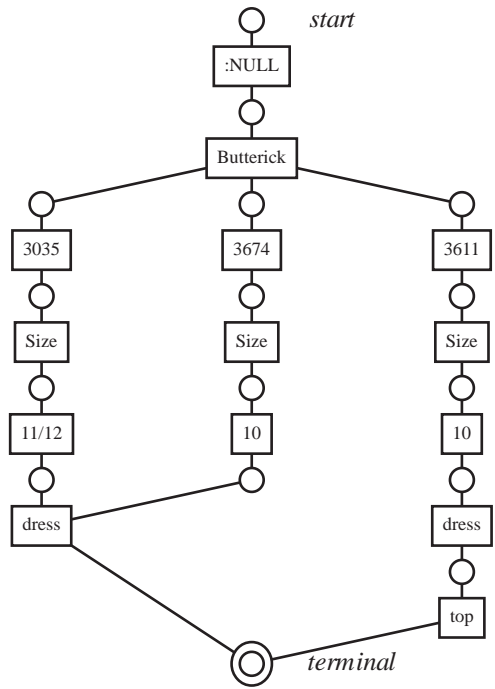

Figure 6: Sample finite-state machine after processing three fabric pattern notes.

Merge any two states if any of the following are true:

1. Another state transitions to both states on the same token; or
   (This enforces determinism.)
2'. Both states have a common 0-leader and
   a. Both states are accepting states, or
   b. Both states transition to a common state via the same token; or
3. Both states have a common 1-leader and
   a. Both states transition to a common state via any token, or
   b. One transitions to the other via any token.

Table 2: Extended FSM state merging rules.

by merging states that share similar transitions. Oddly enough, one motivation for using tree automatons is that they are less likely to introduce extraneous loops, the opposite of the problem with the original FSM merging rules in Table 1. It is not clear how to map the sequence of tokens in the user's notes into a tree structure, but the less sequential nature of the tree automaton may help alleviate sequencing problems in rendering the custom user interface (see Section 9, Observations/Limitations).

### 3.3 Parsing

To use the finite-state machine for prediction, the software needs a strategy for dealing with novel tokens. For example, when the user takes a note about a PowerBook computer with a new memory configuration, the FSM will not have a transition for the first token. If the software is to prompt the user, then it must have a means for deciding where novel tokens lie in a note's syntax—which state to predict from. Without such a mechanism, no meaningful prediction can be generated after novel tokens.

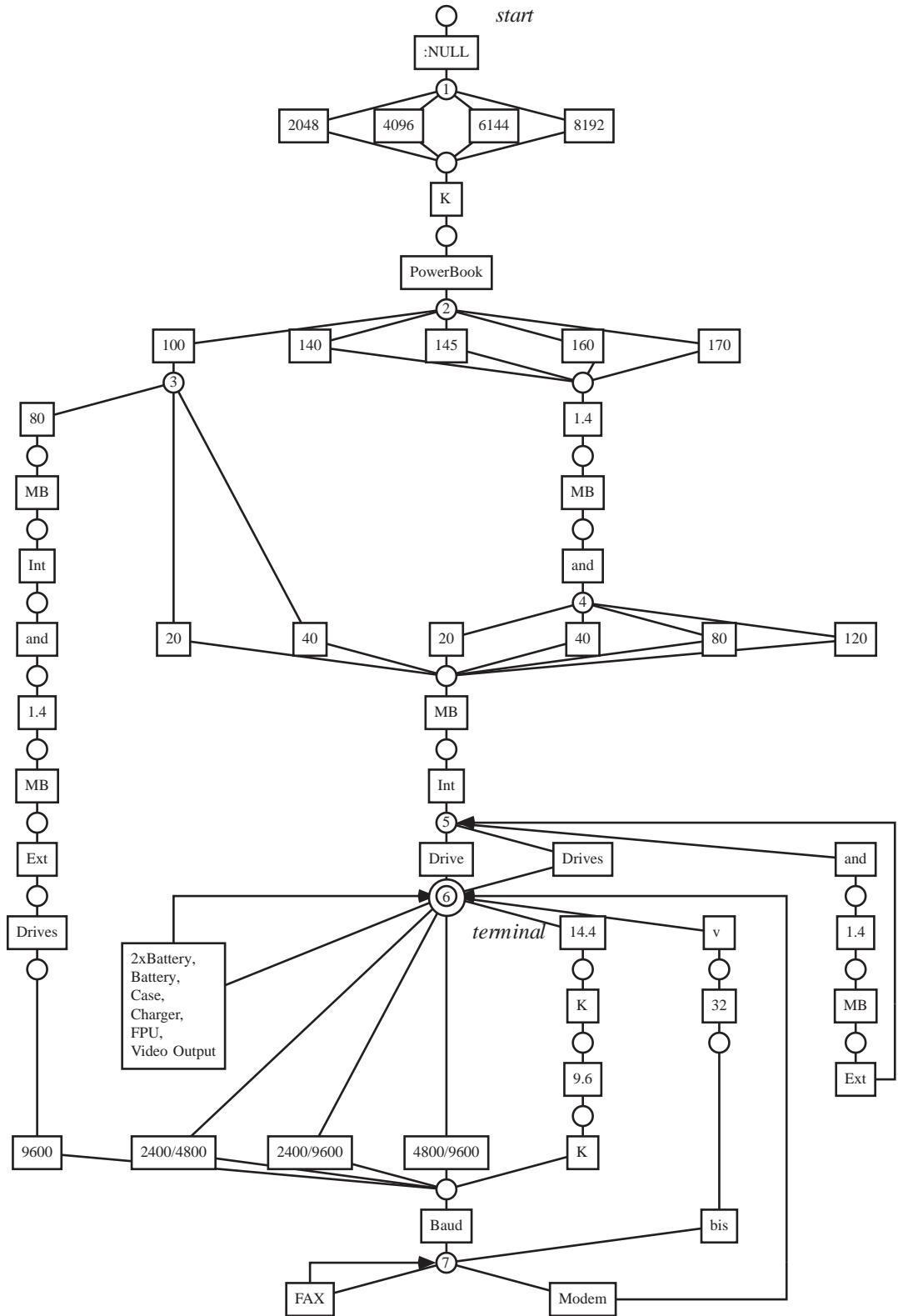

Figure 7: Zero-reversible FSM characterizing PowerBook notes (cf. Example 1).

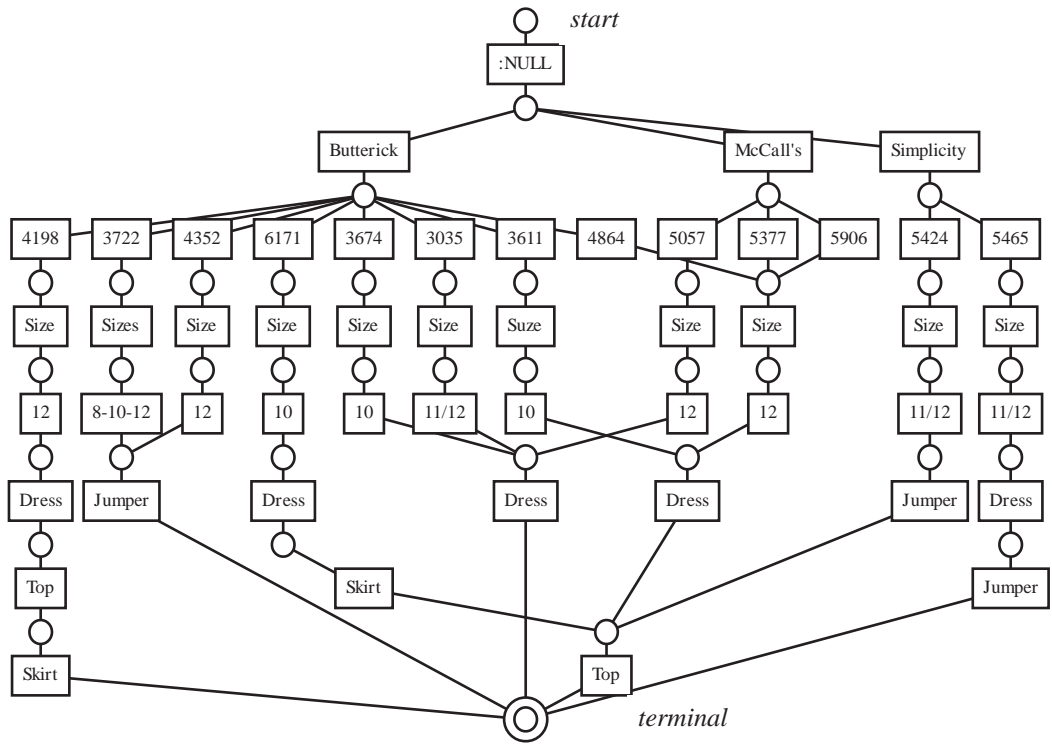

Figure 8: Zero-reversible finite-state machine characterizing fabric pattern notes learned using merging rules listed in Table 1.

A state may not have a transition for the next token. In general, this is a single symptom with three possible causes: (1) a novel token has been inserted, (2) a suitable token has been omitted and the next token would be accepted by a subsequent state, or (3) a token has been simply replaced by another in the syntax. For example, in the sequence of tokens {`:NULL`, `"12288"`, `"K"`, `"PB"`}, `"12288"` is a novel token, a familiar memory size has been omitted, and `"PowerBook"` has been replaced by `"PB"`.

An optimal solution would identify the state requiring a minimum number of insertions, omissions, and replacements necessary to parse the new sequence. An efficient, heuristic approximation does a greedy search using a special marker. Each time the marked state in the FSM has a transition for the next token written by the user, the marker is moved forward, and a prediction is generated from that state. When there is no transition for the next token, a greedy search is conducted for some state (including the marked one and those reachable from it) that has a transition for some token (including the next one and those following). If such a state is found, the marker is moved forward to that state, tokens for the transitions of skipped states are assumed omitted, and novel tokens are assumed inserted. If no state past the marker has a transition for any of the remaining tokens, the remaining tokens are assumed to be replacements for the same number of the most likely transitions; the marker is not moved. If the user writes a subsequent token for which some state has a transition, the marker is moved as described above, and the syntax of the user's note is realigned with the learned syntax. Continuing with the simple PowerBook example, the marker is moved to State 1 of the FSM in Figure 7 because the initial state had a transition for the first token `:NULL`. Because State 1 doesn't have a transition for the next token `"12288"`, a greedy search is conducted to find a nearby state that accepts either `"12288"`, `"K"`, or `"PB"`. The state just before State 2 accepts `"K"`, so the marker is moved to that state. Another greedy search is started to find a state that accepts `"PB"`. Because one cannot be found, the heuristic parsing assumes that it should skip to the next transition. In this case the one labeled `"PowerBook"`. Consequently, the system generates a prediction from State 2 to prompt the user.

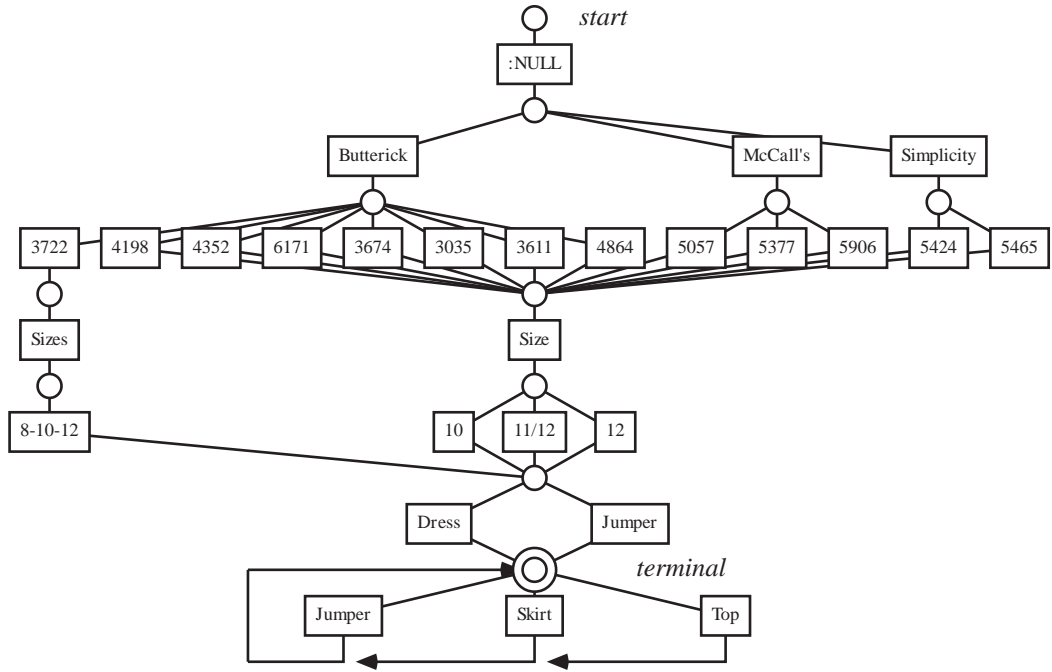

Figure 9: Finite-state machine characterizing fabric pattern notes learned using extended rules in Table 2. Compare to zero-reversible finite-state machine for the same domain in Figure 8.

### 3.4 Multiple Finite-State Machines

If the user decides to take notes about multiple domains, it may be necessary to learn a separate syntax for each domain. For example, a single syntax generalized over both the Power-Book and fabric pattern notes is likely to yield confusing predictions and an unnatural user interface. Maintenance of multiple finite-state machines is an instance of the clustering problem—deciding which notes should be clustered together to share a FSM. As Fisher (1987) discusses, this involves a trade-off between maximizing similarity within a cluster and minimizing similarity between clusters. Without the first criteria, all notes would be put into a single cluster. Without the second criteria, each note would be put into its own cluster.

One obvious approach would be to require the user to prepend each note with a unique token to identify each note's domain. This simplifies the clustering computation. All notes sharing the first token would share a FSM. However, with this scheme, the user would have to remember the identifying token or name for each domain. An interface could provide a pop-up list of all previously used domain identifiers. This is not satisfactory because it requires overhead not needed when taking notes on paper.

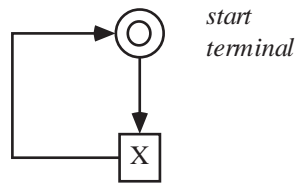

Figure 10: Simple finite-state machine with one state.

An alternative approach doesn't require any extra effort on the part of the user. A new note is grouped with the FSM that skips the fewest of its tokens. This heuristic encourages within cluster similarity because a FSM will accept new token sequences similar to those it summarizes. To inhibit the formation of single-note FSMs, a new FSM is constructed only if all other FSMs skip more than half of the new note's tokens. This is a parametrized solution to encourage between-cluster dissimilarity.

## 4. Learning Embedded Classifiers

Finite-state machines are useful representations for capturing the syntax of a user's notes, and they are easy to learn. When predicting a note's completion, it is essential that a prediction be made from the correct state in the FSM (as discussed above). It is also necessary to decide whether to terminate (indicating acceptance of the note) or continue prediction, and, in the later case, which transition to predict. To facilitate these decisions, the FSM can maintain a count of how many times parsing terminated and how many times each transition was taken. Prediction can then return the option with the maximum frequency.

Figure 10 depicts a FSM for which this method will prove insufficient. There is only one state, an accepting state, and the transition corresponding to the token "x" is optional. (This corresponds to a check box interface item.) There are two problems with a frequency-based prediction. First, the FSM does not indicate that the transition is to be taken at most once, yet this is quite clear from the user interface. Second, simple frequency-based prediction would always recommend termination and never the transition. The FSM accepts whether the box is checked or not, thus the frequency of termination is greater than or equal to the frequency of the transition. This problem arises whenever there is a loop.

Embedding general classifiers in a FSM can alleviate some of the FSM's representational shortcomings. For example, in the FSM depicted in Figure 10, a decision tree embedded in this state easily tests whether the transition has already been taken and can advise against repeating it. Moreover, a classifier can predict based on previous transitions rather than just the frequency of the current state's transitions. Therefore, a decision tree embedded in the state of Figure 10 can predict when the transition should be taken as a function of other, earlier tokens in the sequence. Table 3 lists sample decision trees embedded in states of the FSM depicted in Figure 7. The first tree tests which token was parsed by a distant state, in effect augmenting the FSM representation. It relates memory size to hard disk capacity (small amounts of memory correlate with a small hard disk). The second tree prevents an optional loop from being taken a second time by testing to see if the state has yet been visited during a parse of the note. After processing additional notes, this second decision tree becomes more complex as the system tries to predict which PowerBooks have FAX modems and which do not.

| Decision tree embedded in State 3: | Decision tree embedded in State 7: |
|---|---|
| If State 1 exited with `"2048"` | If State 7 has not been visited |
|     Then predict `" 20"` |     Then predict `" FAX"` |
| Else if with `"4096"` | Else if State 7 exited with `" FAX"` |
|     Then predict `" 40"` | Then predict `" Modem"` . |
| Else if with `"6144"` | |
|     Then predict `" 40"` | |
| Else if with `"8192"` | |
|     Then predict `" 40"` . | |

Table 3: Sample decision trees embedded in the finite-state machine depicted in Figure 7.

A classifier is trained for each state in the FSM which: (a) has more than one transition, or (b) is marked as a terminal state but also has a transition. The classifiers are updated incrementally after the user finishes each note. The classifier's training data are token sequences parsed at this state. The class value of the data is the transition taken from, or termination at, this state by the token sequences. Only those classifiers whose states are used in a parse are updated. The attributes of the data are the names of states prior to this one, and the values of the attributes are the transitions taken from those states. A distinct attribute is defined each time a state is visited during a given parse, so when a loop transition is taken a specific attribute reflects this fact. For any of the attributes, if the corresponding state was not visited while parsing the token sequence, the attribute has a special, empty value.

Consider the PowerBook FSM shown in Figure 7. A classifier would be embedded at States 1, 2, 3, 4, 5, 6, 7. A training example corresponding to the note in Example 1 for the classifier at State 6 would be:

| Attributes: | | Values: |
|---|---|---|
| S1 | = | `"4096"` |
| S2 | = | `" 170"` |
| S3 | = | NIL |
| S4 | = | `" 40"` |
| S5 | = | `" Drives"` |
| S6 | = | `", 2400/9600"` |
| S7 | = | `" FAX"` |
| S7-1 | = | `" Modem"` |
| Class: | = | `:TERMINATE` . |

Note that there is no value for State 3, denoting that it wasn't visited during the parse of Example 1. Also there are two attributes for State 7 denoting that it has been visited twice.

The classifier gives informed advice about which transition to take or whether to terminate. The FSM in turn gives the classifier a specific context for operation. If only a single classifier were used to predict the next token, it would be hard pressed to represent the different predictions required. The domain is naturally narrowed by the FSM and therefore reduces the representational demands on the classifier. Later, we present empirical results comparing a single classifier to a set of classifiers embedded in a FSM. The findings there show that the latter outperforms the former, confirming the intuition that learning is more effective if situated within a narrow context.

From the classifier's point of view, the learning task is non-stationary. The concept to be learned is changing over time because the structure of the FSM is changing. When two states are merged, one of the two classifiers is discarded. The other is now embedded in a different position in the FSM, and it sees different training data. Similarly, when other states are merged, the attributes of the training data also change. To help mitigate this effect, the new state takes the oldest identifier assigned to the two merged states. Empirical results in Table 14 illustrate that the FSM does not have to be fixed before the classifier can learn useful information.

## 5. Contextual Prompting

In the prompting mode, the software continuously predicts a likely completion as the user writes out a note. It presents this as a default next to the completion button. The button's saturation ranges from white to green in proportion to the confidence of the prediction. If the user taps the completion button, the prompt text is inserted at the end of the current note.

A completion is generated by parsing the tokens already written by the user, finding the last state visited in the FSM, and predicting the next most likely transition (or termination). This process is repeated until a stopping criterion is satisfied, which is discussed below. If the last token written by the user is incomplete, matching only a prefix of a state's transition, then the remainder of that transition is predicted. If the last token matches more than one transition, a generalized string is predicted using special characters to indicate the type and number of characters expected. If a digit is expected, a `"#"` is included; if a letter, an `"a"` is included; if either are possible, a `"?"` is included; and if some transition's tokens are longer than others, a `"…"` is appended to the end. For example, if the user has written `"4096K PowerBook 1"`, the possible values for PowerBook models of `"100"`, `"140"`, `"160C"`, and `"170"` are generalized, and the prompt is `"#0…"`.

A simple calculation is used to compute the confidence of the prediction and set the button's color saturation. It is the simple ratio

$$\frac{f(prediction)}{f(total) \times (1 + skipped)}$$

where $f(prediction)$ is the frequency of the predicted arc (or terminate) [i.e., the number of times this choice was taken while parsing previously observed notes], $f(total)$ is the total frequency of all arcs (and terminate), and $skipped$ is the number of tokens skipped during heuristic parsing (cf. Section 3.3, Parsing). Confidence is directly proportional to the simple likelihood of the prediction and is degraded in proportion to the number of tokens the FSM had to skip to get to this point. This information is used in a simple way, so it is unclear if more sophisticated measures are needed.

The stopping criterion is used to determine how much of a prompt to offer the user. At one extreme, only a single token can be predicted. This gives the user little context and may not provide much assistance. At the other extreme, a sequence of tokens that completes the note can be predicted. This may be too lengthy, and the user would have to edit the prompt if selected. The stopping criterion in Table 4 balances these two extremes and attempts to limit prompts to a consistent set of tokens. In particular, Condition 3 stops expanding the prompt

Stop expanding the prompt if any of the following are true:
1. The next prediction is to terminate; or
2. The next prediction is a generalized string; or
3. At least one token has already been predicted and
   a. The prediction starts with punctuation, or
   b. The confidence of the prediction is lower; or
4. The next prediction is the same as the last prediction; or
5. More than 10 tokens have already been predicted.

Table 4: Stopping criterion for contextual prompting.

upon reaching a syntactic boundary (leading punctuation) or upon reaching a semantic boundary (falling confidence).

## 6. Constructing a Button-Box Interface

In the button-box mode, the software presents an interactive graphical interface. Instead of writing out the note, the user may select note fragments by tapping buttons. To switch from contextual mode to button-box mode, a green radio button indicator is displayed below the completion button when the software is confident about the user's syntax. If the user taps this indicator, the existing text is removed, and the corresponding buttons in the button-box interface are selected. As the user selects additional buttons, the interface dynamically expands to reveal additional choices. Because the interface reflects an improving syntactic representation, it also improves with successive notes.

The button-box interface is a direct presentation of a finite-state machine. After the user has written out a token or so of the note, the software finds the FSM that best parses these tokens. The mode switch is presented if the syntax is sufficiently mature—if the average number of times each state has been used to parse earlier notes is greater than 2. If the user selects this indicator, the FSM is incrementally rendered as a set of radio buttons and check boxes.

The two user interface item types correspond to optional choices (check boxes) and exclusive choices (radio buttons). Mapping a FSM into these two item types proceeds one state at a time. Given a particular state to be rendered, any transition that starts a path that does not branch and eventually returns back to the state is rendered as a check box (a loop). The loop corresponds to syntactically optional information. The label for the check box consists of each of the transition labels along the looping path. Other non-looping transitions are rendered as buttons in a single radio button panel along with an extra, unlabeled button. They correspond to syntactically exclusive information. The label for each radio button consists of each transition label up to the point of a subsequent branch or termination. For example, compare the FSM depicted in Figure 7 and the corresponding button-box interface in Figure 3.

Because the transitions for different radio buttons lead to different parts of the FSM, it may confuse the user to render the entire FSM at once. So, each branching state is rendered as it is visited. Initially, the first state in the FSM is rendered. Then, when a radio button is selected, the branching state at the end of its transition path is rendered. Note that check boxes do not trigger additional rendering because the branching state at the end of their loop has already been rendered. This interactive process is repeated as long as the user selects radio buttons that lead to branching states.

## 7. Empirical Results

We tested the interactive note taking software on notes drawn from a variety of domains. Tables 5 through 11 list sample notes from seven domains (in addition to the PowerBook and fabric pattern sample notes listed above).

```
CVA-62 8/6/63 to 3/4/64 Mediterranean A-5A AG 60X

CVA-61 8/5/64 to 5/6/65 Vietnam RA-5C NG 10X
```

Table 5: Sample notes from the airwing domain. Listed above are 2 of the 78 notes about airwing assignments aboard aircraft carriers collected from (Grove & Miller, 1989).

```
B, 81, 5, 151 (2.5), Cyl. 4, 2-bbl., Pontiac

C, 82, X, 173 (2.8), Cyl. 6, 2-bbl., Chevrolet
```

Table 6: Sample notes from the engine code domain. Listed above are 2 of the 20 notes about the meaning of engine codes stamped on automobile identification plates collected from Chilton's Repair & Tune-Up Guide (1985).

```
90, Mazda MPV, 40K MI, 7 Pass, V6, Auto
ABS, PL/PW, Cruise, Dual Air

87, Grand Caravan, 35K MI, 7 Pass, V6, Auto
Cruise, Air, Tilt, Tinting
```

Table 7: Sample notes from the minivan domain. Listed above are 2 of the 22 notes about minivan automobiles collected by the first author.

```
Lorus Disney Oversize Mickey Mouse Watch.
Genuine leather strap.

Seiko Disney Ladies' Minnie Mouse Watch.
Leather strap.
```

Table 8: Sample notes from the watch domain. Listed above are 2 of the 89 notes about personal watches collected from the Best catalog (a department store).

```
azatadine maleate
Blood: thrombocytopenia.
CNS: disturbed coordination, dizziness, drowsiness, sedation,
vertigo.
CV: palpitations, hypotension.
GI: anorexia, dry mouth and throat, nausea, vomiting.
GU: Urinary retention.
Skin: rash, urticaria.
Other: chills, thickening of bronchial secretions.
```

```
brompheniramine maleate
Blood: aganulocytosis, thrombocytopenia.
CNS: dizziness, insomnia, irritability, tremors.
CV: hypotension, palpitations.
GI: anorexia, dry mouth and throat, nausea, vomiting.
GU: urinary retention.
Skin: rash, urticaria.
After parenteral administration:
  local reaction, sweating, syncope may occur.
```

Table 9: Sample notes from the antihistamine domain. Listed above are 2 of the 17 notes on the side effects of antihistamines collected from the Nurses Guide to Drugs (1979).

```
Canon FD f/1.8, 6oz., f/22, 13in.,
good sharpness, poor freedom from flare,
better freedom from distortion,
focal length marked on sides as well as
on front of lens
```

```
Chinon f/1.7, 6oz., f/22, 9in.,
poor sharpness, good freedom from flare,
good freedom from distortion,
cannot be locked in program mode, which
is only a problem, of course, when lens is
used on program-mode cameras
```

Table 10: Sample notes from the lens domain. Listed above are 2 of the 31 notes about 35mm SLR camera normal lenses collected from the Consumer Reports (1988).

---

```
22in. W. 48in.
A very large falcon. Three color phases occur:
blackish, white, and gray-brown. All
are more uniformly colored than the
Peregrine Falcon, which has dark
mustaches and hood.
```

---

```
16-24in. W. 42in.
Long-winged, long-tailed hawk with a
white rump, usually seen soaring
unsteadily over marshes with its wings
held in a shallow 'V'. Male has a pale
gray back, head, and breast. Female
and young are brown above, streaked
below, young birds with a rusty tone.
```

---

Table 11: Sample notes from the raptor domain. Listed above are 2 of the 21 notes about North American birds of prey collected from (Bull & Farrand, 1977).

Summary characteristics of the nine domains are listed in Table 12 together with some simple measures to indicate prediction difficulty. For instance, Column 1 shows the number of notes in the domain. With a larger number of notes, the easier it should be to accurately train a predictive method. Column 4 shows the standard deviation (STD) of the length of all notes in each domain. It is more likely that a well-behaved FSM can be discovered when STD is low. In this and successive tables, the domains are ranked by STD. Column 5 presents the percentage of unique tokens in the notes. The fewer novel tokens a note has, the more likely that successive tokens can be predicted. This measure places an upper bound on predictive accuracy. Column 6 shows the percentage of constant tokens, ones that always appear in a fixed position. It is easier to predict these constant tokens. Finally, Column 7 indicates the percentage of repeated tokens. When fewer tokens are repeated verbatim within a note, the more likely that the predictive method will not become confused about its locale within a note during prediction.

The first six domains are natural for the interactive note taking task because they exhibit a regular syntax. The last three domains are included to test the software's ability on less suitable domains. Notes from the Antihistamine, Lens, and Raptor domains contain highly-variable lists of terms or natural language sentences. Learned FSMs for notes in these domains are unlikely to converge, and, in the experiments reported here, only the FSM for the Lens data exceeded the maturity threshold (average state usage greater than 2).

## 7.1 Contextual Prediction Accuracy

Column 7 of Table 13 lists the accuracy of next-token predictions made by the software in prompting mode. The first nine rows list predictive accuracy over all tokens as notes from each of the nine domains are independently processed in the order they were collected. The last row lists predictive accuracy over all tokens as notes from all nine domains are collectively processed. This simulates a user taking notes about several domains simultaneously.

To put these results in context, the table also lists predictive accuracies for several other methods. Column 1 lists the accuracy for a lower bound method. It assumes that each note shares a fixed sequence of tokens. Termed *common*, this method initializes its structure to the

| Domain | 1<br>N Notes | 2<br>N Tokens | 3<br>Tokens/Note | 4<br>STD | 5<br>% Unique | 6<br>% Constant | 7<br>% Repeated |
|---|---|---|---|---|---|---|---|
| Airwing | 78 | 936 | 12.0 | 0.3 | 18 | 8 | 0 |
| Pattern | 13 | 75 | 5.8 | 0.7 | 21 | 0 | 0 |
| Engine Code | 20 | 222 | 11.1 | 0.8 | 0 | 0 | 0 |
| Minivan | 22 | 335 | 15.2 | 1.7 | 9 | 17 | 0 |
| PowerBook | 95 | 1238 | 13.0 | 2.6 | 1 | 31 | 15 |
| Watch | 89 | 832 | 9.3 | 5.1 | 13 | 0 | 1 |
| Antihistamine | 17 | 421 | 24.8 | 9.4 | 17 | 8 | 1 |
| Lens | 31 | 1066 | 34.4 | 9.6 | 1 | 26 | 19 |
| Raptor | 21 | 878 | 41.8 | 11.5 | 33 | 7 | 22 |

Table 12: Quantitative properties of the nine domains used to test alternative methods.

first note. It then removes each token in this sequential structure that cannot be found in order in other notes. At best, this method can only predict the constant, delimiter-like tokens that may appear regularly in notes. Its performance is limited by the percentage of constant tokens reported in Column 6 of Table 12. It performs best for the PowerBook notes where it learns the following note syntax:

    * :NULL * "K" * " PowerBook" * "MB" * "MB" * " Int." * .    (Example 3)

(The asterisks indicate Kleene star notation.) This reads as some sequence of zero or more tokens then the token `:NULL`, followed by zero or more tokens then `"K"`, followed by zero or more tokens then `"PowerBook"`, and so on. It is less successful for the minivan notes where it learns a simpler syntax:

    * :NULL * "K" * " MI" * " Pass" * .    (Example 4)

Columns 2 and 3 of Table 13 list the accuracy of using a classifier to directly predict the next token without explicitly learning a syntax. In this paradigm, examples are prefixes of token sequences. Attributes are the last token in the sequence, the second to last token, the third to last token, and so on. Class values are the next token in the sequence—the one to be predicted. Column 2 lists the performance of a simple Bayes classifier, and Column 3 lists the performance of an incremental variant of ID3 (Schlimmer & Fisher, 1986). Perhaps surprisingly, these methods perform considerably worse than the simple conjunctive method. Without the benefit of a narrow context provided by the FSM, these methods must implicitly construct representations to detect differences between similar situations that arise within a single note. For example, in the PowerBook notes, a classifier-only approach must learn to discriminate between the first and second occurrence of the `"MB"` token.

Column 4 of Table 13 lists the accuracy of a more viable prediction mechanism. Based on simple ideas of memorization and termed *digram*, the method maintains a list of tokens that have immediately followed each observed token. For example, in the fabric pattern domain, this method retains the list of tokens {`"8-10-12"`, `"10"`, `"11/12"`, `"12"`} as those that follow the token `"Size"`. Each list of *follow* tokens are kept in order from most to least frequent. To predict the next token, the system looks for the last token written and predicts the most frequent follow token. This method is nearly as effective as any other in Table 13, especially on the combined task when notes from each domain are entered in random order. Laird (1992) describes an efficient algorithm for maintaining higher-dimensional n-grams, in effect increasing the context of each prediction and effectively memorizing longer sequences of tokens. Laird's algorithm builds a Markov tree and incorporates heuristics that keep the size of the tree from growing excessively large. Regrettably, these methods are unsuitable for the interactive note-taking software because of the difficulty of using them to construct a custom user interface. It is plausible to construct a panel of exclusive choices based directly on the set of follow tokens, but it is unclear how to identify optional choices corresponding to loops in finite-state machines. Moreover, if notes are drawn from different domains, and those domains share even a single token, then some follow set will include tokens from different domains. Using these follow sets to construct a user interface will unnecessarily confuse the user by introducing options from more than one domain at a time.

| Domain | 1 Common | 2 Bayes | 3 ID4 | 4 Digram | 5 FSM | 6 FSM+Bayes | 7 FSM+ID4 | 8 Upper |
|---|---|---|---|---|---|---|---|---|
| Airwing | 19 | 8 | 8 | 47 | 62 | 44 | 62 | 79 |
| Pattern | 25 | 15 | 16 | 34 | 43 | 43 | 51 | 68 |
| Engine Code | 18 | 8 | 8 | 59 | 64 | 63 | 69 | 87 |
| Minivan | 29 | 6 | 7 | 54 | 46 | 44 | 47 | 80 |
| PowerBook | 40 | 7 | 8 | 73 | 70 | 76 | 82 | 96 |
| Watch | 21 | 10 | 14 | 44 | 39 | 33 | 42 | 78 |
| Antihistamine | 11 | 4 | 6 | 40 | 24 | 22 | 24 | 68 |
| Lens | 22 | 3 | 3 | 68 | 63 | 60 | 63 | 91 |
| Raptor | 9 | 2 | 3 | 11 | 12 | 9 | 12 | 55 |
| Combined | — | — | — | 48 | 46 | 45 | 49 | — |

Table 13: Percentage of tokens correctly predicted as a function of the learning method.

Column 5 of Table 13 lists the accuracy of prediction based solely on the learned FSMs. Without an embedded classifier, this method must rely on prediction of the most common transition (or termination) from each state. Because the prediction is based on simple counts (as noted in Section 4, Learning Embedded Classifiers), this method *never* predicts optional transitions.

Columns 6 and 7 of Table 13 list the accuracy of predicting using FSMs and embedded classifiers. The classifiers used are simple Bayes and the incremental ID3, respectively. The latter outperforms either the FSM alone or the FSM with embedded Bayes classifiers. If the system only makes predictions when its confidence measure is greater than 0.25, the accuracy is significantly different for the Engine Code, Minivan, Lens, and Raptor domains, ranging between 10 and 22 percentage points of improvement.

Column 8 of Table 13 lists an estimate of the upper-bound on predictive accuracy. This was calculated by assuming that prediction errors were only made the first time each distinct token was written.

| Domain | 1<br>Norm | 2<br>Diff<br>Tokens | 3<br>Rules<br>2a,b | 4<br>Rules<br>2ab,3a | 5<br>No<br>Restart | 6<br>Accept<br>= 1/4 | 7<br>Accept<br>= 3/4 | 8<br>Repeat<br>Atts | 9<br>Drop<br>Class'r | 10<br>New IDs |
|---|---|---|---|---|---|---|---|---|---|---|
| Airwing | 62 | 62 | 63 | 62 | 62 | 62 | 62 | 62 | 61 | 63 |
| Pattern | 51 | 51 | 53 | 52 | 50 | 51 | 51 | 51 | 51 | 53 |
| Engine Code | 69 | 71 | 72 | 69 | 43 | 69 | 69 | 69 | 67 | 72 |
| Minivan | 47 | 48 | 48 | 47 | 28 | 47 | 47 | 52 | 45 | 48 |
| PowerBook | 82 | 80 | 83 | 83 | 77 | 82 | 82 | 81 | 80 | 82 |
| Watch | 42 | 42 | 43 | 43 | 28 | 42 | 42 | 42 | 41 | 43 |
| Antihistamine | 24 | 25 | 24 | 24 | 9 | 24 | 24 | 24 | 24 | 24 |
| Lens | 63 | 66 | 64 | 63 | 46 | 63 | 63 | 63 | 63 | 64 |
| Raptor | 12 | 11 | 12 | 12 | 11 | 12 | 12 | 12 | 12 | 12 |

Table 14: Percentage of tokens correctly predicted as a function of design variations.

## 7.2 Design Decisions

The note taking software embodies a number of design decisions. Table 14 lists the effects of these decisions on predictive accuracy by comparing versions of the software with and without each design feature. The first column lists the predictive accuracy for the software's nominal configuration. Column 2 lists the accuracy data for a slightly different generic tokenizer. Accuracy is higher for some domains, lower for others. A custom-built tokenizer is one way to incorporate knowledge about the domain. Columns 3 and 4 show the accuracy for the system using only the original two FSM merging rules (cf. Table 1) and all but the last merging rule (cf. Table 2), respectively. The decreased structural generality tends to lower predictive accuracy, but the embedded classifiers help compensate for the reduced accuracy. Column 5 lists the accuracy for when the FSM does not heuristically continue parsing upon encountering a token for which there is no immediate transition. As expected, accuracy suffers considerably in some domains because a novel token in a sequence completely foils any subsequent prediction. Columns 6 and 7 list accuracy for different values of the free parameter controlling the clustering of notes together into a FSM. There is little effect on predictive accuracy in this case. Column 8 shows the accuracy for when embedded classifiers do not use information about repeated states in the FSM. Without this information, the classifiers cannot predict that a loop transition should be taken exactly once. Surprisingly, elimination of this feature has little effect on accuracy. Column 9 lists the accuracy for when the embedded classifiers associated with a pair of FSM states are discarded when the states are merged. Finally, Column 10 lists the accuracy for when a new FSM state is assigned a unique ID rather than the ID of the oldest of the two merged states.

## 7.3 Sample Button-Box Interfaces

In addition to Figure 3, Figures 11 through 15 depict button-box interfaces for the five other well-behaved note taking domains listed at the top of Table 12. These interfaces are visual and offer the user an organized view of their notes, presenting options in a natural way. However, whenever unique tokens are involved, the current software makes no attempt to explicitly generalize tokens. This effect is reflected in the tour dates for the Airwing notes in Figure 11. Note that the radio button panel consists of a long series of dates, none of which is likely to be selected for a new note.

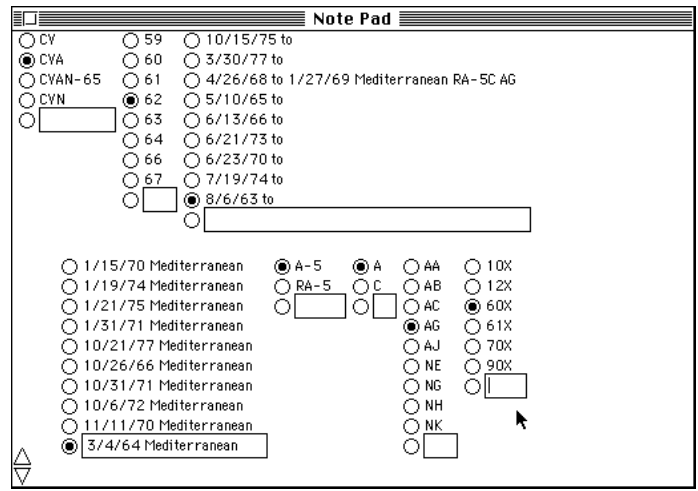

Figure 11: Screen snapshot of the note-taking software in button-box mode for an airwing note.

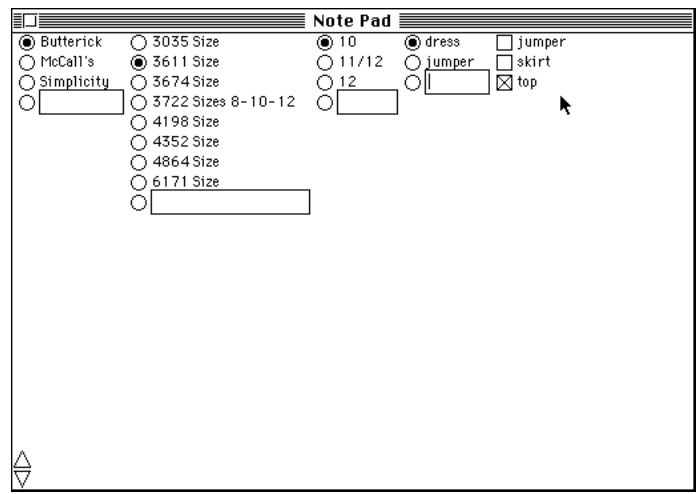

Figure 12: Screen snapshot of the note-taking software in button-box mode for a fabric pattern note.

## 8. Related Work

Self-customizing software agents have several subjective dimensions on which they can be evaluated and compared:

- *Anticipation*—Does the system present alternatives without the user having to request them?
- *User interface*—Is the system graphical, or is it command-line oriented?
- *User control*—Can the user override or choose to ignore predictive actions?
- *Modality*—If the system has a number of working modes, can the user work in any mode without explicitly selecting one of them?
- *Learning update*—Is learning incremental, continuous and/or real-time?

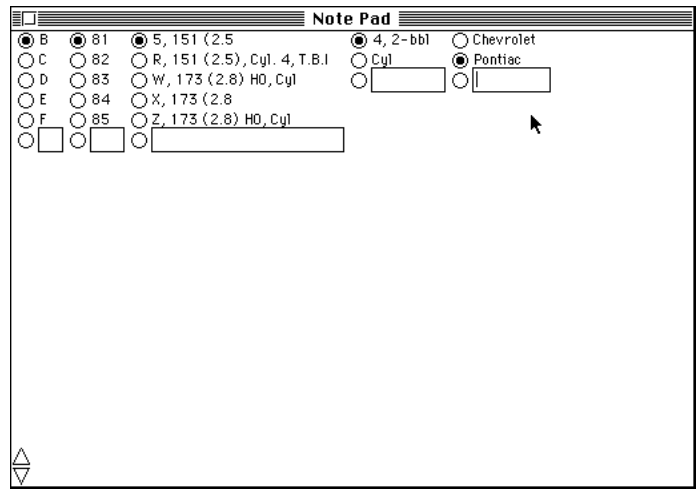

Figure 13: Screen snapshot of the note-taking software in button-box mode for an engine code note.

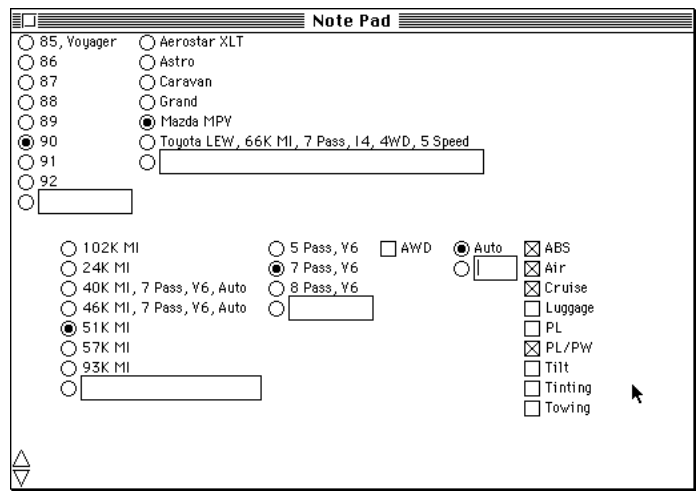

Figure 14: Screen snapshot of the note-taking software in button-box mode for a minivan note.

- *User adjustable*—Can the user tune the system parameters manually?

Here we describe related systems that exhibit properties in each of these agent dimensions.

Our note taking software utilizes the *anticipation* user interface technique pioneered by Eager (Cypher, 1991). Eager is a non-intrusive system that learns to perform iterative procedures by watching the user. As such, it is a learning apprentice, a software agent, and an example of programming by example or demonstration. Situated within the HyperCard environment, it continuously watches a user's actions. When it detects the second cycle of an iteration, it presents an execute icon for the user's notice. It also visually indicates the anticipated next action by highlighting the appropriate button, menu item, or text selection in green. As the user performs their task, they can verify that Eager has learned the correct procedure by comparing its anticipations to their actions. When the user is confident enough, they can click on the execution icon, and Eager will run the iterative procedure to completion. Eager is highly anticipatory, uses a graphical interface, is non-obtrusive, non-modal, and learns in real-time, but is not user adjustable.

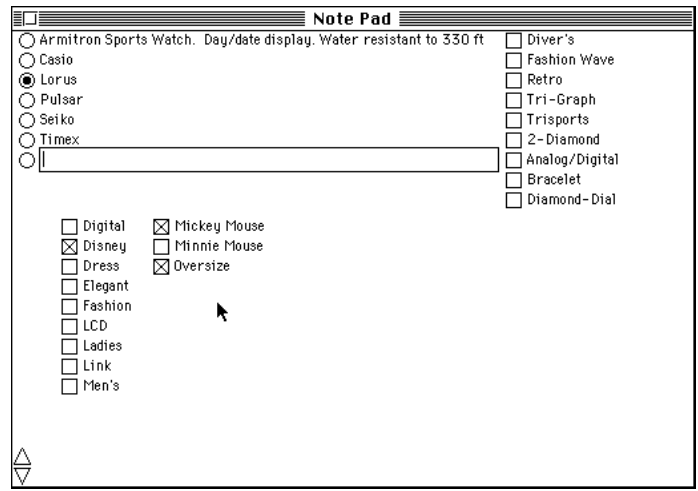

Figure 15: Screen snapshot of the note-taking software in button-box mode for a watch note.

CAP is an apprenticeship system that learns to predict default values (Dent, et al., 1992). Its domain of operation is calendar management, and it learns preferences as a knowledgable secretary might. For example, a professor may prefer to hold a regular group meeting in a particular room at a particular time of day for a particular duration—information that a secretary would know from experience. CAP collects information as the user manages their calendar, learns from previous meetings, and uses the regularities it learns to offer default values for meeting location, time, and duration. The learning system is re-run each night on the most recent meeting data, and the learned rules are applied for prediction the following day. CAP is also designed to utilize an extensible knowledge base that contains calendar information and a database of personnel information. The system continues to be used to manage individual faculty calendars. Though offering some intelligence, CAP's user interface is line-oriented and is based on the Emacs editor. Questions asked of the user about meetings are presented using a command-line dialog, and the default predictions are displayed one-at-a-time. CAP can be characterized as anticipatory, command-line oriented and modal with user control (but not user adjustable), where learning is done in batch.

Another related system addresses the task of learning to fill out a form (Hermens & Schlimmer, 1993). The system recreates a paper form as an on-screen facsimile, allowing the user to view all of the pertinent information at a glance. Input typed by the user into the electronic form is processed by a central form-filling module. When the user completes a form copy, it is printed, and each field value on the form is forwarded to a learning module (a decision tree learning method). The learned representations predict default values for each field on the form by referring to values observed on other fields and on the previous form copy. From the user's point of view, it is as if spreadsheet functions have been learned for each field of the form. Empirical studies indicate that this system reduced the number of key-strokes required of the user by 87% on 269 forms processed over the 8 month period in which it was actually used by office personnel. This system is unobtrusive, non-modal and anticipatory, uses a graphical interface, and updates learning in real-time.

Maes and Kozierok (1993) are addressing the problem of self-customizing software at a much more task-independent level. They identify three learning opportunities for a software agent: observing the user's actions and imitating them, receiving user feedback upon error, and incorporating explicit training by the user. To illustrate the generality of their framework, they demonstrate simple learning apprentices that help sort the user's electronic mail and schedule meetings. Their initial systems use an instance-based (case- or memory-based) approach primarily because it allows efficient update and because it naturally generates a confidence in each of its predictions. User's may set thresholds on these predictions, corresponding to a minimum confidence for when the agent should prompt the user (a "tell-me" threshold) and a higher minimum confidence for the agent to act immediately on behalf of the user (a "do-it" threshold). The framework for learning in this case is anticipatory, utilizes a graphical user interface, is devoted to user control, is non-modal, learns in real-time, and is user adjustable.

A system developed for Macintosh Common Lisp (MCL) provides a word-completion mechanism for word prefixes typed by the user in any window. J. Salem and A. Ruttenberg (unpublished) have devised MCL methods to display a word completion in the status bar of the each window. If the user desires to add the completion to the window, they simply press the `CLEAR` key. This word completion mechanism is similar to file-name completion in EMACS and the C-shell in UNIX systems, except that the word is displayed for the user before it is added. This system is anticipatory (unlike the UNIX file completion), is command line oriented (but displays the default completion in a graphical window), can be fully controlled by the user, is non-modal, learns in real time, is not intended to be user adjustable (though knowledgeable MCL programmers could easily make changes to the code).

The interactive note taking software we have devised does not require any user programming. It only receives implicit user feedback when the user chooses to complete a note in a different way than prompted. It does not have any mechanisms for direct user instruction or threshold tuning. In a system designed to be as easy to use as paper, such explicit adjustment may be inappropriate. We characterize our system as anticipatory, graphically-oriented, and modal (due to the switching that takes place when a user wishes to display the button-box interface). It allows the user to override default prompts and predictions, and it learns in real-time. We have not included features that allow the user to configure the performance of the agent.

## 9. Observations/Limitations

The interactive note-taking software is designed to help users capture information digitally, both to speed entry and improve accuracy, and to support the longer term goal of efficient retrieval. The software incorporates two distinctive features. First, it actively predicts what the user is going to write. Second, it automatically constructs a custom radio-button, checkbox user interface.

This research explores the extremes of FSM learning and prediction, where the system has no explicit *a priori* knowledge of the note domains. We have tried to design the system so that it can learn quickly, yet adapt well to semantic and syntactic changes, all without a knowledge store from which to draw. It is clear that knowledge in the form of a domain-specific tokenizer would aid FSM learning by chunking significant phrases and relating similar notations and abbreviations. Some preliminary work has shown that, after a few notes have been written, users may create abbreviations instead of writing out whole words. A domain-specific tokenizer would be able to relate an abbreviation and a whole word as being in the same class, and therefore allow for more flexibility during note taking. For example, a domain-specific tokenizer may recognize that `"Megabytes"`, `"Meg"`, `"MB"`, and `"M"` all represent the same token for memory sizes. One could imagine a framework that would allow for domain-specific tokenizers to be simply plugged in.

The prototype built to demonstrate these ideas was implemented on a conventional, micro computer with keyboard input. As a consequence, it was impossible to evaluate user acceptance of the new interface or the adaptive agent. With newly available computing devices incorporating pen input and handwriting recognition, it should be possible to re-engineer the user interface and field test these ideas with actual users.

One aspect of note learning, related to tokenization and the button-box user interface display, is the difficulty of generalizing numeric strings or unique tokens. The cardinality of the range of model numbers, telephone numbers, quantities, sizes, other numeric values, and even proper names is very large in some note domains. The finite-state machine learning method presented here is incapable of generalizing over transitions from a particular state, and, as a consequence, the current system has the problem of displaying a very lengthy button-box interface list. (A button is displayed for each value encountered in the syntax of notes, and there may be many choices.) For example, a large variety of pattern numbers may be available in the fabric pattern note domain. An appropriate mechanism is desired to determine when the list of numeric choices is too large to be useful as a button-box interface. The system can then generalize the expected number, indicating the number of digits to prompt the user: `####`, for example. This may be helpful to remind the user that a number is expected without presenting an overbearing list of possibilities.

Another limitation of the current effort lies in the choice of finite-state machines to represent the syntax of the user's notes. Notes may not be regular expressions with the consequence that the FSMs may become too large as the learning method attempts to acquire a syntax. This may place an unreasonable demand on memory and lead to reduced prompting effectiveness.

The choice of finite-state machines also apparently constraints the custom user interface. Because FSMs branch in unpredicable ways, button-box interfaces must be rendered incrementally. After the user indicates a particular transition (by selecting a button), the system can render states reachable from that transition for the user. Ideally, the user should be able to select buttons corresponding to note fragments in any order, allowing them to write down the size before the pattern number, for example. To construct a non-modal user interface, a more flexible syntactic representation is needed.

Several of the low-level design decisions employed in this system are crude responses to technical issues. For instance, the decision to render a syntax as a button-box interface only if the average number of times each state has been used to parse notes is greater than 2. This ignores the fact that some parts of the state machine have been used frequently for parsing notes while other parts have rarely been used. Similarly, the particular measure for estimating prompting confidence (and setting the saturation of the completion button) is simplistic and would benefit from a more sound statistical basis.

## Acknowledgments

Anonymous reviewers suggested an additional example in Section 3, offered some refinements to the user interface, graciously identified some limitations of the work listed in Section 9, and pointed out some additional related work. Mike Kibler, Karl Hakimian, and the EECS staff provided a consistent and reliable computing environment. Apple Cambridge developed and supports the Macintosh Common Lisp programming environment. Allen Cypher provided the tokenizer code. This work was supported in part by the National Science Foundation under grant number 92-1290 and by a grant from Digital Equipment Corporation.

## References

Angluin, D. (1982). Inference of reversible languages. *Journal of the Association for Computing Machinery*, *29*, 741–765.

Berwick, R. C., & Pilato, S. (1987). Learning syntax by automata induction. *Machine Learning*, *2*, 9–38.

Bull, J., & Farrand, J., Jr. (1977). The Audubon Society Field Guide to North American Birds (Eastern Edition). NY: Alfred A. Knopf (pp. 401–682).

*Chilton's Repair & Tune-Up Guide: GM X-Body 1980-1985* (1985). Randor, PA: Chilton Book (p. 7).

Cohen, W. W. (1988). Generalizing number and learning from multiple examples in explanation based learning. *Proceedings of the Fifth International Conference on Machine Learning* (pp. 256–269). Ann Arbor, MI: Morgan Kaufmann.

Consumer Reports (1988), *53* (12), 302–303. Mount Vernon, NY: Consumers Union.

Cypher, A. (1991). Eager: Programming repetitive tasks by example. *Proceedings of CHI* (pp. 33–39). New Orleans, LA: ACM.

Dent, L., Boticario, J., McDermott, J., Mitchell, T., & Zabowski, D. (1992). A personal learning apprentice. *Proceedings of the Tenth National Conference on Artificial Intelligence* (pp. 96–103). San Jose, CA: AAAI Press.

Fisher, D. H. (1987). Knowledge acquisition via incremental conceptual clustering. *Machine Learning*, *2*, 139–172.

Grove, M., & Miller, J. (1989). North American Rockwell A3J/A-5 Vigilante. Arlington, TX: Aerofax (pp. 13–15).

Hermens, L. A., & Schlimmer, J. C. (1993). A machine-learning apprentice for the completion of repetitive forms. *Proceedings of the Ninth IEEE Conference on Artificial Intelligence for Applications*. Orlando, FL.

Laird, P. (1992). Discrete sequence prediction and its applications. *Proceedings of the Tenth National Conference on Artificial Intelligence* (pp. 135–140). San Jose, CA: AAAI Press.

Maes, P., & Kozierok, R. (1993). Learning interface agents. *Proceedings of the Eleventh National Conference on Artificial Intelligence* (pp. 459–465). Washington, D. C.: AAAI Press.

Nurse's Guide to Drugs (1979). Horsham, PA: Intermed Communications (pp. 454–462).

Schlimmer, J. C., & Fisher, D. H. (1986). A case study of incremental concept induction. *Proceedings of the Fifth National Conference on Artificial Intelligence* (pp. 496–501). Philadelphia, PA: AAAI Press.